\documentclass[10pt,english]{article}
\usepackage{subfigure}
\usepackage{graphicx}
\usepackage{float}
\usepackage[T1]{fontenc}
\usepackage[latin9]{inputenc}
\usepackage[margin=1.15in]{geometry}
\usepackage{url}
\usepackage{float}
\usepackage{amsmath}
\usepackage{amsbsy}
\usepackage{setspace}
\usepackage{amssymb}
\usepackage{esint}
\usepackage{cite}
\newfloat{algorithm}{tbp}{loa}
\floatname{algorithm}{Algorithm}
\usepackage{algorithmic}

\newlength\myindent
\makeatletter

\floatstyle{ruled}
\newfloat{algorithm}{tbp}{loa}
\floatname{algorithm}{Algorithm}

\usepackage{babel}

\begin{document}

\title{Additive Feature Hashing}

\author{M. Andrecut}


\maketitle
{

\centering Calgary, Alberta, T3G 5Y8, Canada

\centering mircea.andrecut@gmail.com

} 
\bigskip 
\begin{abstract}
The hashing trick is a machine learning technique used to encode categorical features into a numerical vector representation of pre-defined fixed length. 
It works by using the categorical hash values as vector indices, and updating the vector values at those indices. 
Here we discuss a different approach based on additive-hashing and the "almost orthogonal" property of high-dimensional random vectors. 
That is, we show that additive feature hashing can be performed directly by adding the hash values and converting them into high-dimensional 
numerical vectors. We show that the performance of additive feature hashing is similar to the hashing trick, and we illustrate the 
results numerically using synthetic, language recognition, and SMS spam detection data. 

\smallskip 

\end{abstract}

\section{Introduction}

The Bag-of-Words (BOW) model is a widely used model in natural language processing and information retrieval. 
In the BOW model a document is represented by a high-dimensional vector, with the components corresponding to the frequency of a particular token (word) in the document \cite{key-1}. 
These vectors can then be used for training machine learning algorithms for different tasks, like document classification. 
In this case, the BOW model requires a vocabulary of all unique words occurring in all the documents in the training set. 
Such a vocabulary can grow very fast, for example the English vocabulary size is estimated to be between $10^4$ and $10^5$ words, 
creating difficulties in implementing the machine learning algorithms, since the BOW vectors will have a very high dimensionality. Such problems are usually mitigated by employing 
some dimensionality reduction techniques, prior to feeding the vectors to the machine learning algorithms. However, the high memory storage requirement for a large vocabulary 
still remains, even after reducing the vectors dimensionality. 

A more efficient solution is the so called "hashing trick" (HT), where tokens are mapped directly to vector indices using a hashing function \cite{key-2,key-3}. 
The HT approach works by transforming the token hash values into vector indices, and then updating the vector values at those indices, such that no memory is required to store the large vocabulary. 
Thus, the HT simplifies the implementation of BOW models and improves scalability of machine learning methods.  

Inspired by the HT model, here we discuss a different approach based on additive-hashing (AH) and the "almost orthogonal" property of high-dimensional random vectors \cite{key-4,key-5,key-6}. 
That is, we show that additive feature hashing can be performed directly by adding the hash values and converting them into high-dimensional 
numerical vectors. We show that the performance of additive feature hashing is similar to the hashing trick, and we illustrate the 
results numerically using synthetic, language recognition, and SMS spam detection data. Moreover, we show that this simple approach 
does not even require a "learning" process in order to achieve similar results to more "expensive" machine learning methods. 

\section{The Hashing Trick}

Let us give simple example by considering the following three documents:

\begin{enumerate}
\item John likes to watch movies
\item Mary also likes to watch movies
\item Jane makes popcorn
\end{enumerate}
The vocabulary corresponding to these documents is: John: 1; likes: 2; to: 3; watch: 4; movies: 5; Mary: 6; also: 7; Jane: 8; makes: 9; popcorn: 10. 
Using this vocabulary the documents can be represented by the following term-document matrix:

\begin{equation}
\begin{bmatrix}
    John & likes & to & watch & movies & Mary & also & Jane & makes & popcorn \\
    1 & 1 & 1 & 1 & 1 & 0 & 0 & 0 & 0 & 0 \\
    0 & 1 & 1 & 1 & 1 & 1 & 1 & 0 & 0 & 0 \\
    0 & 0 & 0 & 0 & 0 & 0 & 0 & 1 & 1 & 1
\end{bmatrix}
\end{equation}

Each row in the matrix corresponds to the BOW vector representation of a document. The problem with the BOW model 
is that such document matrices will grow in size as the training set grows, requiring a large amount of storage space. 

In order to circumvent this dimensionality growth problem one can use the HT method, where instead of creating 
a term-document matrix, that maps words to indices, one can simply use a hash function that hashes the 
words directly to indices \cite{key-2,key-3}. That is we consider a hash function $h:\{strings\} \rightarrow [0,1,...,L-1]$ that 
maps any given string into a range of integers. Thus the hash value of any word will point directly 
to an index in an $L$-dimensional vector, and then one can simply update the vector at that index, and thus creating 
a $L$-dimensional vector representation of any document. 
The pseudo-code implementation of this process is shown in Algorithm 1. 

Almost any hash function can be used in this process, with the only requirement of being approximately uniform \cite{key-2,key-3}. 
Since most hash functions will return a large integer, exceeding the imposed value of $L$, one can use the modulo 
operator to limit the output into the $[0,1,...,L-1]$ range. This, however will generate hash collisions. 
That is, the hashing of different words will point to the same index. It has been shown that in order to counter 
the effect of hash collisions one can use a second single-bit hash function $s:\{strings\} \rightarrow \{-1,+1\}$ 
to determine the sign of the update value in the vector representation, as shown in Algorithm 2.

\begin{algorithm}[t!]
\caption{The hashing trick.}
\begin{algorithmic}[1]
\STATE $\textbf{function} \text{ hashing-trick}(document,L)$
\STATE $v \leftarrow$ zeros$(L)$
\FOR{$word \textbf{ in } document$}
	\STATE $i \leftarrow h(word) \text{ mod } L$
	\STATE $v[i] \leftarrow v[i] + 1$
\ENDFOR
\RETURN $v$
\end{algorithmic}
\end{algorithm}

\begin{algorithm}[t!]
\caption{The hashing trick, collision mitigation.}
\begin{algorithmic}[1]
\STATE $\textbf{function} \text{ hashing-trick-cm}(document,L)$
\STATE $v \leftarrow$ zeros$(L)$
\FOR{$word \textbf{ in } document$}
	\STATE $i \leftarrow h(word) \text{ mod } L$
	\STATE $v[i] \leftarrow v[i] + s(word)$
\ENDFOR
\RETURN $v$
\end{algorithmic}
\end{algorithm}

The HT method is conveniently implemented in the HashingVectorizer class of the sklearn python module \cite{key-7}. The hash function used in this implementation is the signed 32-bit version of Murmurhash3. 
One can easily use this class to convert the three example documents into vectors of size $L=32$, and to compute the similarity between the documents using the 
dot product between the corresponding vectors, as shown below:

\begin{verbatim}
import numpy as np
from sklearn.feature_extraction.text import HashingVectorizer

if __name__=="__main__":
    d = ["John likes to watch movies",
        "Mary also likes to watch movies",
        "Jane makes popcorn"]
    hvec = HashingVectorizer(n_features=32)
    v = hvec.transform(d).toarray()
    print("similarity(d[0],d[1]):",np.dot(v[0],v[1]))
    print("similarity(d[0],d[2]):",np.dot(v[0],v[2]))
    print("similarity(d[1],d[2]):",np.dot(v[1],v[2]))
>
similarity(d0,d1)= 0.7302967433402215
similarity(d0,d2)= 0.0
similarity(d1,d2)= 0.0
\end{verbatim}

The feature vectors are L2 normalized, and therefore their dot product is equal to the cosine of the angle between the vectors, 
such that the more parallel the vectors are, the more similar the documents are, and vice versa. One can see that the first two 
documents are quite similar, while the third is not similar to the first two.

\section{Additive Hashing}

Instead of using the hash values of the tokens as indices, what about adding the token hash values and then converting them into high-dimensional 
numerical vectors? Here we show that this approach is also feasible, and it is based on incremental hash functions \cite{key-8,key-9}, and the "almost orthogonal" property of high-dimensional random vectors \cite{key-4,key-5,key-6}. 

Consider the strings $x,x' \in \{ strings \} $, such that $x'$ is a modification of $x$. 
A function $f: \{ string \} \rightarrow \lbrace 0,1 \rbrace^L$ is said to be incremental if one can update $f(x')$ in 
time proportional to the amount of modification between $x$ and $x'$ rather than having to recompute $f(x')$ from scratch.

The incremental hash functions can be constructed using The Randomize-then-Combine Paradigm for collision-free hash functions, containing \cite{key-6}:
(1) A randomizer function $h(.)$ that maps the bit strings to elements of a group;
(2) A combine operator $\odot$ corresponding to the group operation of these elements.

A function $h: \{ strings \} \rightarrow G$ that maps strings to an abelian group $G$ is called a randomizer function.
Assume that the output of the randomizer function has a fixed length $L$, and the string $x$ is the concatenation ($\parallel$) of $n$ tokens:
\begin{equation}
x = x_1 \parallel x_2 \parallel... \parallel x_n,
\end{equation}
then the randomizer function $h(.)$ maps these tokens to the group elements as following:
\begin{equation}
g_i = h(x_i) \in G, \forall i=1,...,n.
\end{equation}
The group operation $\odot$ of a randomizer function $h:\{ strings \} \rightarrow (G,\odot)$ is called the combining operator. 

Therefore, the incremental hash functions can be defined as following \cite{key-9}. 
Assume that $(G,\odot)$ is an abelian group, and $h:\{ strings \} \rightarrow G$ is a randomizer function, then the function $H_h^G: \{ strings \} \rightarrow G$ is called an incremental hash function if the hash value of $x = x_1 \parallel x_2 \parallel... \parallel x_n$ is:

\begin{equation}
H^G_h(x) = \bigodot_{i=1}^n h(x_i).
\end{equation}

If the token $x_i$ is changed to $x'_i$, then the whole string $x$ changes as:
\begin{equation}
x' = x_1 \parallel x_2 \parallel ... \parallel x_{i-1} \parallel x'_i \parallel x_{i+1} \parallel ... \parallel x_n,
\end{equation}
however using the incremental property, the hash of $x'$ can be easily computed from the previous hash value of $x$ as following:
\begin{equation}
H^G_h(x') = H^G_h(x)\odot h^{-1}(x_i) \odot h(x'_i),
\end{equation}
where $h^{-1}(x_i) \in G$ is the inverse of the group element $h(x_i) \in G$.

The randomizer function $h(.)$ is considered to be an ideal random oracle, however in practice it can be
derived from a standard hash functions like the SHA family. In this conditions, the security of the incremental hash function 
relies only on the security of the combining operation $\odot$, which is typically a computationally hard problem in a chosen group. 
Different levels of security can be attained for incremental hash functions, since one can use different randomizer functions and groups. 
Bellare and Micciancio (BM) have proposed three incremental hash functions that can be proven secure \cite{key-9}.
Since we are mostly interested in applying the incremental hash to the BOW vector encoding problem, here we only consider the 
additive hash function, which uses modular addition as a combining operation in  $G = \mathbb{Z}_N$: 
\begin{equation}
H^G_h(x) = \sum_{i=1}^n h(x_i)\:(\text{mod}\: N).
\end{equation}
One can show that this hash function is collision-free as long as the randomizer function $h$ is ideal, and the 
Weighted Knapsack Problem (WKP) is a hard problem. The WKP can be formulated as following:
for a $L$-bit positive integer $N$ and $q$ numbers $a_1, a_2,..., a_q \in \mathbb{Z}_N$, find weights $w_1,w_2,...,w_q \in \lbrace -1, 0, 1 \rbrace$ not all zero, such that:
\begin{equation}
\sum_{i=1}^q w_i a_i \:(\text{mod}\:N)=0.
\end{equation}

In our case, the BOW vector encoding problem, we are are not interested in very strong collision free guarantees, so we can relax this requirement, 
and for a practical implementation we only require that the randomizer function $h$ output has a large enough number of bits $L$. 
Moreover, for each token $x_i$, or sub-string, of $x = x_1 \parallel x_2 \parallel... \parallel x_n$ we consider the vector: 
\begin{equation}
v^{(i)} = \frac{1}{\sqrt{L}} [v^{(i)}_0,v^{(i)}_1,...,v^{(i)}_{L-1}]^T, \: \Vert v^{(i)} \Vert = 1,
\end{equation}
with the elements $v^{(i)}_\ell=2h_\ell(x_i)-1$, where $h_\ell$ is the $\ell$-th output bit of the randomizer function $h(x_i)$. 
We also replace the group operation $\odot$ with the vector addition in the linear vector space. 
Thus, the string $x=x_1 \parallel x_2 \parallel... \parallel x_n$ will be represented by the vector sum:
\begin{equation}
v = \sum_{i=1}^n v^{(i)}.
\end{equation}
In order to be able to use the dot product as a similarity measure we L2 normalize the vector as following:
\begin{equation}
v \leftarrow v/\Vert v \Vert.
\end{equation}
With above equations we practically perform the BOW encoding process, from tokens to vectors:
\begin{align}
x_i \rightarrow h(x_i) \rightarrow v^{(i)}, \\
x \rightarrow \sum_{i=1}^n h(x_i) \rightarrow \sum_{i=1}^n v^{(i)} = v
\end{align}

Since the token vectors $v^{(i)}$ are constructed from the output of a random function, the expectation value of 
the dot product for any two vectors corresponding to two different tokens is zero: 
\begin{equation}
E(v{(i)}^T v^{(j)}) = \frac{1}{L} E \left( \sum_{\ell=1}^L v^{(i)}_\ell v^{(j)}_\ell \right) = 0, 
\end{equation}
while the expected value of the standard deviation is: $\sigma=1/\sqrt{L}$. Thus, the vectors corresponding 
to any two different tokens are "almost orthogonal" if $L$ is large enough and there are no collisions. 

A practical implementation of the additive hashing (AH) can be also easily done in python using the shake\_256() function from the hashlib module \cite{key-10}, as shown below:

\begin{verbatim}
import numpy as np
from hashlib import shake_256

def randomizer(w,L):
    h = shake_256(w.encode()).digest(L//8)
    b = bin(int.from_bytes(h,byteorder="little"))[2:].zfill(L)
    v = 2*np.array(list(map(int,b))) - 1
    return v/np.sqrt(L)

def ahash(d,L):
    v = np.zeros((len(d),L))
    for i in range(len(d)):
        for w in d[i].split(" "):
            v[i] = v[i] + randomizer(w,L)
        v[i] = v[i]/np.linalg.norm(v[i])
    return v

if __name__=="__main__":
    d = ["John likes to watch movies",
        "Mary also likes to watch movies",
        "Jane makes popcorn"]
    v = ahash(d,32)
    print("similarity(d[0],d[1]):",np.dot(v[0],v[1]))
    print("similarity(d[0],d[2]):",np.dot(v[0],v[2]))
    print("similarity(d[1],d[2]):",np.dot(v[1],v[2]))

>
similarity(d[0],d[1]): 0.7778061881946695
similarity(d[0],d[2]): -0.1737020834449128
similarity(d[1],d[2]): -0.25833561143518957
\end{verbatim}

The shake\_256() function provides variable length digests with up to 256 bits of security. As such, the digest method requires a length, and the maximum digest length is not limited \cite{key-10}. 
The randomizer() function takes a token w, computes the shake\_256 digest of length $L/8$ bytes, then a binary array of length $L$ is created from the digest, and finally the binary array 
is transformed into a normalized vector. The ahash() function performs the additive feature hashing by iterating over all the words in a list of documents, and calling the randomizer() 
function to encode the tokens into vectors, which are then added together and the result is L2 normalized. One can see that qualitatively the result is similar to HT, 
the first two documents are similar, while the third is not similar to the first two.

\section{Application to natural language processing tasks}

\subsection{Synthetic Data}

In order to compare the AH and HT methods we consider the following synthetic data case. 
Assume that $s$ is a string of length $M=100$, randomly generated from the set of lowercase letter characters. 
We consider the task of measuring the similarity between this string $s$ and an altered version of it $s'$. 
The alteration is done gradually, such that each character in the original string $s$ is randomly changed 
to another ascii character in $s'$ with a probability $p \in [0,1]$. Then the similarity $\eta(s,s',p)$ is calculated 
as a function of $p$, by using all the n-grams of length $n$ as the BOW model extracted from the strings. 
The similarity should decrease from $\eta(s,s',0)=1$, when $p=0$, to an expectation of $\eta(s,s',1)=0$, 
when $p=1$. For each value of $p\in [0,1]$ we consider $T=100$ strings and we compute the average of $\eta(s,s'p)$ 
for different vectors dimensionality: $L=2^\ell$. 
The numerical results show equivalent results for both AH and HT methods, for all n-gram lengths and dimensionality $L$. 
In Figure 1 we show the results for n-grams of length $n=3$, and $L=2^\ell$, $\ell=7,8,9,10$. 

\begin{figure}[t!]
\centering \includegraphics[width=16cm]{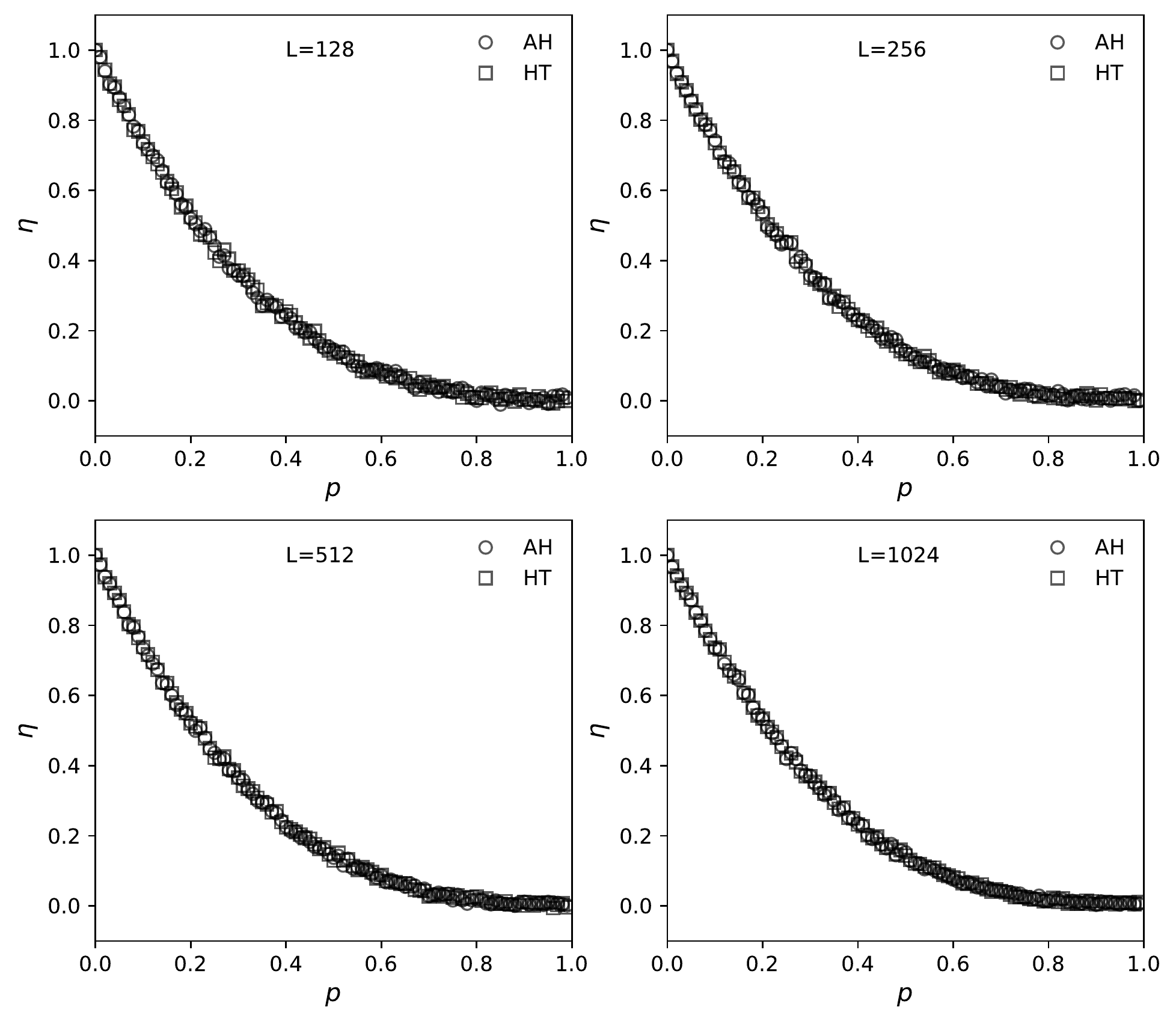}
\caption{AH and HT synthetic string similarity results for n-grams of length $n=3$, and vector dimensionality $L=128,256,512,1024$.}
\end{figure}

\subsection{Language Recognition}

In the second task we compare the performance of the AH and HT methods in language recognition. 
In order to do so we use the WiLI-2018 benchmark dataset \cite{key-11}, which was created by extracting paragraphs from Wikipedia. 
It total there are 235,000 paragraphs extracted for 235 languages, 1000 paragraphs per language. 
The language classes include: 122 Indo-European languages, 22 Austronesian languages, 17 Turkic languages, 14 Uralic languages, 11 Niger-Congo languages, 10 Sino-Tibetan languages, 9 Afro-Asiatic
languages, 6 constructed languages and 24 languages of smaller families. 
The mean paragraph length averaged over all languages is 371 unicode characters, the shortest paragraph is guaranteed to be 140 unicode characters long, while the longest paragraph is 195,402 unicode characters.
Each language also contains short excerpts from other languages and special characters.

The WiLI-2018 dataset is a text classification dataset, that is given an unknown paragraph, one has to decide in which language it is written. 
The dataset is divided into a training set and testing set. Both sets contain 117,500 paragraphs from the 235 languages, with 500 paragraphs per language. 
In \cite{key-11} a neural network solution implemented in Keras and Tensorflow, with a tf-idf feature vectorizer in sklearn (2,218 features), is described. 
The neural network is a multilayer perceptron with one hidden layer of 512 neurons, followed by the ReLU activation function, and an output layer
of 235 neurons followed by softmax is used as a classifier. The network has 1,256,683 parameters, and the cross-entropy loss 
function with the Adam optimizer was used to train it over 20 epochs, with a mini-batch size of 32. The paper reports that the neural network achieved an accuracy of $88.30\%$ \cite{key-11}.
Here we show that both AH and HT give better results just by simply using the cosine similarity, and the nearest neighbor (highest dot product value) as a classification method, without any learning involved. 

For both AH and HT we use the n-gram approach and we extract the corresponding feature vectors, L2 normalized. We let the length $L=2^\ell$ of the 
feature vector to vary for $\ell=4,5,...,12$. 
The best classification results, with an accuracy of $92.16\%$, were obtained for n-grams of lengths $n=3$, and $\ell=11,12$, that is a length of $L\geq 2,048$ for the feature vectors. 
Moreover, both AH and HT show very close, almost indistinguishable results (Figure 2). 

\begin{figure}[t!]
\centering \includegraphics[width=10cm]{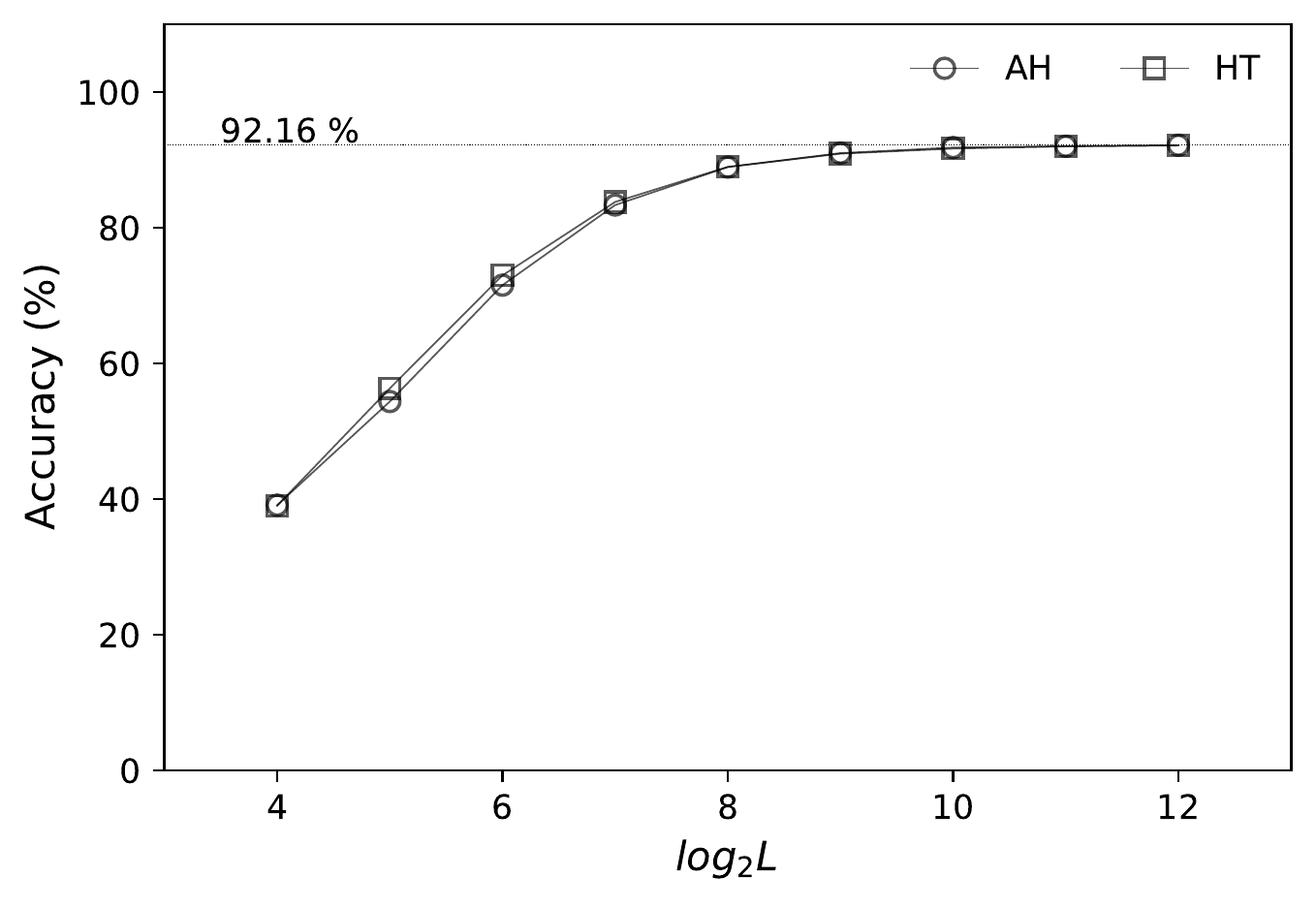}
\caption{AH and HT language recognition results for n-grams of length $n=3$, and vector dimensionality $L=2^\ell$, $\ell=4,5,...,12$.}
\end{figure}

\subsection{SMS Spam Filtering}

The role of spam filters is to protect users from spam and phishing messages. 
Most spam filters are word-based filters, which simply block any email that contains certain words or phrases. 
More sophisticated filters are built around machine learning classification algorithms, which are trained on large sets of already classified spam and non-spam messages. 

Here we apply the AT and HT methods to an SMS spam dataset \cite{key-12}. The dataset contains 5,574 SMS messages, from which 747 are spam ($13\%$), and 4,827 ($86.6\%$) are ham (non-spam). 
The task is to classify unknown SMS messages as ham or spam. 
In \cite{key-12} several machine learning methods have been implemented and used for classification. From the 17 methods used there, the best reported results were 
for the Support Vector Machines classifier, as following: $97.64\%$ Accuracy (ACC), $83.1\%$ Spam Caught (SC), and $0.18\%$ Blocked Hams (BH) \cite{key-12}. 

Using the previously n-gram approach we convert the messages into feature vectors using the AH and HT methods, and we apply the 
same dot product as a similarity measure, with the nearest neighbor (highest dot product value) as a classification method, again without any learning involved. We let the number of features vary as $L=2^\ell$, with $\ell=4,5,...,13$. 
For each value of $L$ we randomly divide the dataset into training and testing sets ($50\%$ each) for $T=100$ times and we 
compute the average ACC, SC and BH values. The results for n-grams of length $n=3$ are given in Figure 3. One can see that for $\ell=12,13$ both AH and HT achieve: 
ACC=$97.41\%$ accuracy, SC=$87.5\%$, and BH=$1.05\%$. These results are very close to the best machine learning results reported in [12], even though 
no learning was employed in our approach. 

\begin{figure}[t!]
\centering \includegraphics[width=8cm]{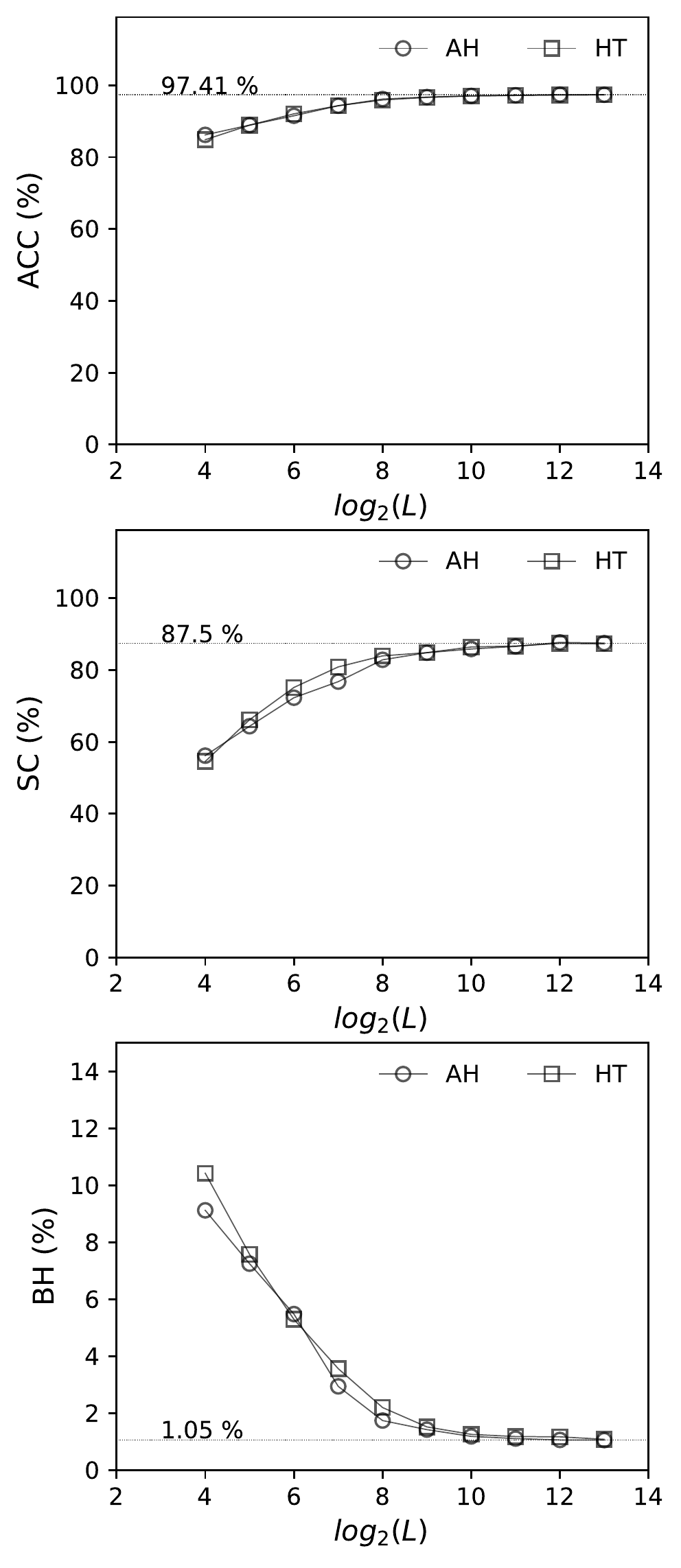}
\caption{AH and HT results for SMS spam filtering, for n-grams of length $n=3$, and vector dimensionality $L=2^\ell$, $\ell=4,5,...,13$ (ACC=Accuracy, SC=Spam Caught, BH=Blocked Hams).}
\end{figure}

\section{Conclusion}

In this paper we have described a new additive-hashing method for text feature extraction, based on incremental hashing functions and 
the "almost orthogonal" property of high-dimensional random vectors. 
We have shown that additive feature hashing can be performed directly by adding the hash values and converting them into high-dimensional 
numerical vectors. Also, we shown that the performance of additive feature hashing is similar to the hashing trick, and we illustrated the 
results numerically using synthetic, language recognition, and SMS spam detection data. Moreover, we have shown that this simple approach 
does not even require a "learning" process in order to achieve similar (or better) results to the more "expensive" machine learning methods.

\end{document}